# Comparing the Performance of L*A*B* and HSV Color Spaces with Respect to Color Image Segmentation

Dibya Jyoti Bora[1], Anil Kumar Gupta[2], Fayaz Ahmad Khan[3]

[1,2,3]Department of Computer Science & Applications, Barkatullah University, Bhopal, India

*Abstract*— Color image segmentation is a very emerging topic for image processing research. Since it has the ability to present the result in a way that is much more close to the human eyes perceive, so today's more research is going on this area. Choosing a proper color space is a very important issue for color image segmentation process. Generally L*A*B* and HSV are the two frequently chosen color spaces. In this paper a comparative analysis is performed between these two color spaces with respect to color image segmentation. For measuring their performance, we consider the parameters: mse and psnr . It is found that HSV color space is performing better than L*A*B*.

## I. INTRODUCTION

Image Segmentation is the most challenging phase of image processing research. This is challenging in the sense that depending upon the result of the image segmentation, the progress of the whole image analysis process depends. In general terms, image segmentation can be defined as the problem of the partitioning a given image into a number of homogeneous segments, such that the union of any two neighboring segments yields a heterogeneous segment [1]. These partitions are homogeneous with respect to some criteria like color, texture, motion etc. Images may be either gray scale or color images. So, image segmentation may be either gray scale image segmentation or color image segmentation. Color image segmentation is totally different from gray scale image segmentation, e.g., content based image retrieval [2][3].Also, segmentation can be either local or global [4]. Local segmentation is small windows on a whole image and deals with segmenting sub image. Global segmentation deals with segmenting whole image. Global segmentation mostly deals with relatively large no of pixel. But local segmentation deals with lower no of pixels as compare to global segmentation [4].There exists different methodologies for the image segmentation process.

In broad terms, these methodologies can be classified into seven main groups [4][5][6]: (1) Histogram thresholding, (2) Clustering (Soft and Hard), (3) Region growing, region splitting and merging, (4) Edge-based, (5) Physical model based, (6) Fuzzy approaches, and (7) Neural network and GA (Genetic algorithm) based approaches. Clustering methodology is found to be very efficient while dealing with image segmentation [4]. Color image segmentation is more utilizable than gray scale image segmentation because of its capability to enhance the image analysis process thereby improving the segmentation result. But when we consider color image segmentation, then choosing a proper color space becomes the most important issue [7]. L*A*B* and HSV are the two frequently chosen color spaces [8][9]. In this paper, we have performed a comparative analysis between these two color spaces to analyze their performance with respect to color image segmentation. This comparative research is done considering "noise" as the major factor. Often the presence of noise in the result of the image segmentation degrades the progress of image analysis process. So, we should optimize our image segmentation process in such a way that noise should be as low as possible in the final segmented image. By considering this issue, we have taken "MSE" and "PSNR" are the measuring criteria for the mentioned performance analysis task. Both of these two are related to the noise measurement in an image. We have proposed a new approach for color image segmentation in[8]. So, we adopted this approach for our comparative study of the color spaces. This paper is designed as follows: first of all a review on the existing work is given. Then we have performed a study on "color image segmentation" and "color spaces : L*A*B* & HSV". After that we have presented the flowchart of the followed approach. We have illustrated each topic involved in the mentioned approach for color image segmentation shown in the flowchart. Then we come to the experimental section followed by the analysis of the result obtained. Finally the conclusion section is given along with the future research enhancement issues.

## II. Review Of Literature

In [9], the authors analyzed the properties of the hsv color space with emphasis on the visual perception of the variation in Hue, Saturation and Intensity values of an image pixel. They extracted pixel features by either choosing the Hue or the Intensity as the dominant property based on the Saturation value of a pixel. The segmentation using this method gives better identification of objects in an image compared to those generated using RGB color space.

In [10], the authors proposed a color image segmentation approach where they first converted the rgb image into hsv one. Then they applied Otsu's multi-thresholding on V-channel to get the best threshold from the image. The resultant image is then segmented with K-Means clustering to merge the over segmented regions that occurred due to the application of Otsu's multi-thresholding. Finally they performed background subtraction along with morphological processing. The result of the approach is found to be quite satisfactory as per the values of MSE and PSNR obained from the experiment.

In [11], the authors first converted the original image from RGB form to HSV form. Then they applied mean shift and FELICM separately on Hue, Saturation and Value Components. The final images obtained from mean shift and FELICM is fused together. The proposed method show better performance level than the previous algorithms.

In [12], the authors proposed a new quantization technique for HSV color space to generate a color histogram and a gray histogram for K-Means clustering, which operates across different dimensions in HSV color space. In this approach, the initialization of centroids and the number of clusters are automatically estimated. A filter for post-processing is introduced to effectively eliminate small spatial regions. This method is found to achieve high computational speed and the results are closed to human perceptions. Also with this method, it becomes possible to extract salient regions of images effectively.

In [13], authors transformed the image from rgb color space to l*a*b* color space. After separating three channels of l*a*b*, a single channel is selected based on the color under consideration. Then, genetic algorithm is applied on that single channel image. The method is found to be very effective in segmenting complex background images.

In [14], the image is converted from rgb to l*a*b* color space, and then k means algorithm is applied to the resulting image. Then pixels are labeled on the segmented image. Finally images are created which segment the original image by color.

In [15], a novel approach for clustering based color image segmentation is proposed. Here, l*a*b* color space is chosen. Then a combined effort of K-Means algorithm, sobel filter and watershed algorithm is used for the segmentation task. The result is found to be quite satisfactory in terms of the MSE and PSNR values.

In [16], the authors proposed a robust clustering algorithm where clustering is performed on the l*a*b* color space. Image segmentation is straightforwardly obtained by setting each pixel with its corresponding cluster. The algorithm is applied for medical image segmentation and the experimental results clearly show region of interest object segmentation.

In [17], the authors proposed an ant based clustering technique with respect to CIE Lab color space where CMC distance is used to calculate the distance between pixels as this distance measure is found to be producing good results with respect to the CIE Lab color space. The performance of this technique is compared with MSE parameter and found to be satisfying one.

## III. Color Image Segmentation

An image is a meaningful arrangement of regions and objects. Image analysis is the process of extracting information from an image which is one of the preliminary steps in pattern recognition systems. Image segmentation can be defined as the classification of all the picture elements or pixels in an image into different clusters that exhibit similar features [4]. The first step in image analysis is to segment the image. This segmentation subdivides an image into its constituent parts or objects to a particular level. This level of subdivision depends on the problem being viewed. Sometimes we need to segment the object from the background to understand the image correctly and identify the content of the image. For this reason, we have mainly two techniques for segmentation: (1) discontinuity detection technique and (2) similarity detection technique. In the first one, the common approach is to partition an image based on abrupt changes in gray-level image. While in the second technique is based on the threshold and region growing [18]. The second one is the most common approach for color image segmentation. Color image segmentation can be defined as a process of extracting from the image domain one or more connected regions satisfying uniformity (homogeneity) criterion which is based on feature(s) derived from spectral components. These components are defined in chosen color space model [19]. So, color space plays the vital rule in color image segmentation.

Color image segmentation is useful in many applications like multimedia applications, text extraction from a color image, skin tumor feature identification, segmentation of colored topographic maps, initial segmentation for knowledge indexing etc[19].

## IV. COLOR SPACES

A color space is actually a specific organization of colors which in combination with physical device profiling, allows us for reproducible representation of colors in both digital and analog representations [20]. This is a special tool for understanding the color capabilities of a particular device or digital file. When we try to reproduce color on another device, then this is the color space which can show us whether we will be able to retain shadow or highlight detail, color saturation and by how much either can be compromised[21].

We know usually color can be measured by the following attributes [22]:

*(1) Brightness:* This is the human sensation by which an area exhibits more or less light.

*(2) Hue:* This is the human sensation according to which an area appears to be similar to one, or to proportions of two, of the perceived colors red, yellow, green and blue.

*(3) Colorfulness:* This is the human sensation according to which an area appears to exhibit more or less of its hue.

*(4) Lightness:* This can be defined as the sensation of an area's brightness relative to a reference white in the scene.

*(5) Chroma:* This is the colorfulness of an area relative to the brightness of a reference white.

*(6) Saturation:* This is the colorfulness of an area relative to its brightness.

According to the tri-chromatic theory, the three separate lights red, green and blue can match any visible color based on the eye's use of three color sensitive sensors. It is the way that most computer color spaces operate, using three parameters to define a color. Thus a color can be described using three co-ordinates or parameters which describe the position of the color within the color space being used. But these co-ordinates do not tell us what color is this as it is totally dependent on the color space being used [22].

Although RGB is the most commonly preferred one, there exists a variety of color spaces for use. This is because different color spaces present color information in different ways that make certain calculations more convenient and also provide a way to identify colors that is more intuitive. E.g., RGB color space defines a color as the percentage of red, green and blue hues mixed together, while others describe colors by their hue(green), saturation(dark green), and luminance or intensity[23]. Again the color spaces can be classified as device dependent and device independent color spaces. A device dependent color space is a color space where the color produced depends on both the parameters used and on the equipment used for display. For example try specifying the same RGB values on two different workstations, the color produced will be visually different if viewed on side by side screens [22]. Means RGB is a device dependent color space. A device independent color space is the one where a set of parameters will produce the same color on whatever equipment they are used. E.g., L*A*B* is a device independent color space. Since this paper is concerned with L*A*B* and HSV color spaces, so in the following subsections, we will discuss a few lines about these two color spaces.

*(A) L*A*B* Color Space:*

This color space is originally defined by CAE and specified by the International Commission on Illumination [24][25]. In this color space, we have one channel is for Luminance (Lightness) and other two color channels are a and b known as chromaticity layers. The a* layer indicates where the color falls along the red green axis, and b* layer indicates where the color falls along the blue-yellow axis. a* negative values indicate green while positive values indicate magenta; and b* negative values indicate blue and positive values indicate yellow. Most important feature of this color space is that this is device independent [26], means to say that this provides us the opportunity to communicate different colors across different devices. Following figure clearly illustrates the coordinate system of l*a*b* color space [27]:

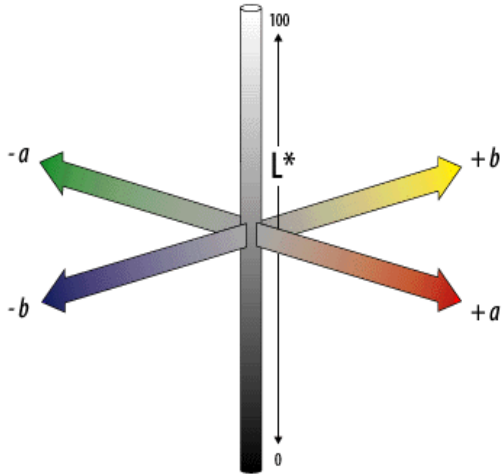

**Fig (I): L*A*B* Color Space**

In the above figure, the central vertical axis represents lightness (L*). L* can take values from 0(black) to 100(white). The co ordinate axes follows the fact that "a color cannot be both red or green, or both blue and yellow, because these colors oppose each other". Values run from positive to negative for each axis. Positive 'a' values indicate amounts of red, while negative values indicate amounts of green. And, positive 'b' values indicate amount of yellow while, negative 'b' values indicate blue. The zero represents neutral gray for both the axes. So, here, values are needed only for two color axes and for the lightness or grayscale axis (L*).

Following figures show a RGB image (obtained from Berkeley Segmentation Dataset [28]) and its respective L*A*B* converted image:

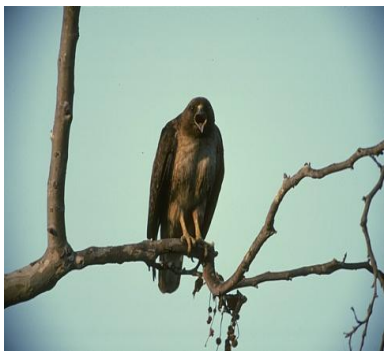

**Fig (II): Original Image**

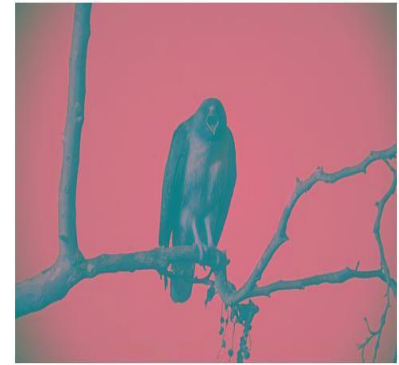

**Fig (III): L*A*B* Converted Image of the Original Image**

*(B) HSV Color Space:*

We can represent HSV color space with the help of a hexacone in three dimensions where the central vertical axis represents the intensity [29]. Here 'H' stands for 'Hue'. The "Hue" is an angle in the range $[0,2\pi]$ relative to the red axis with red at angle 0, green at $2\pi/3$, blue at $4\pi/3$ and red again at $2\pi$[30][31]. 'S' stands for 'Saturation', which describes how pure the hue is with respect to a white reference. This can be thought of as the depth or purity of color and is measured as a radial distance from the central axis with values between 0 at the center to 1 at the outer surface. For S=0, as one moves higher along the intensity axis, one goes from black to white through various shades of gray. While, for a given intensity and hue, if the saturation is changed from 0 to 1, the perceived color changes from a shade of gray to the most pure form of the color represented by its hue[31]. 'V' stands for 'Value' which is a percentage goes from 0 to 100. This range (from 0 to 100) can be thought as the amount of light illuminating a color [32]. For example, when the hue is red and the value is high, the color looks bright. On the other hand, when the value is low, it looks dark. A Diagrammatic view of the HSV color space is found in [33]:

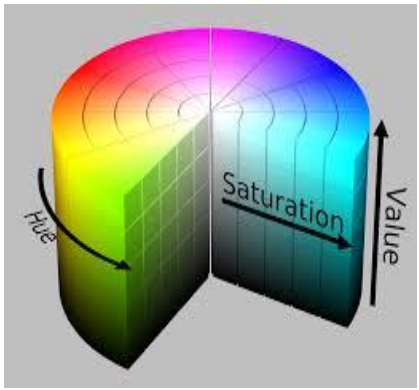

**Fig (IV): HSV Color Space**

Following figures show the HSV converted image of the original RGB image [28]:

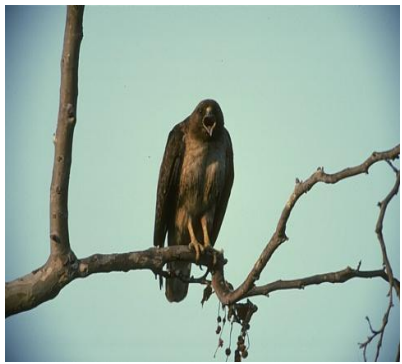

**Fig (V): Original Image**

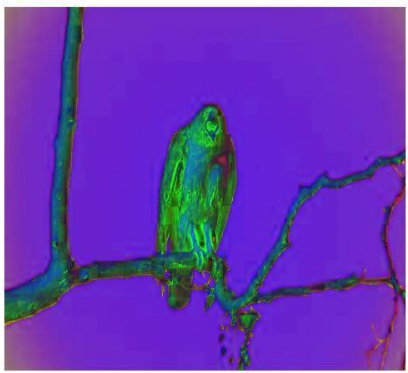

**Fig (VI): HSV Converted Image of the Original Image**

### V. FLOWCHART OF THE APPROACH FOLLOWED

In [8], we have proposed a new approach for color image segmentation and the result of this approach is found to be quite satisfactory.

So, for the comparative study between L*A*B* color space and HSV color space, we have adopted this approach. The flowchart of this approach is shown below:

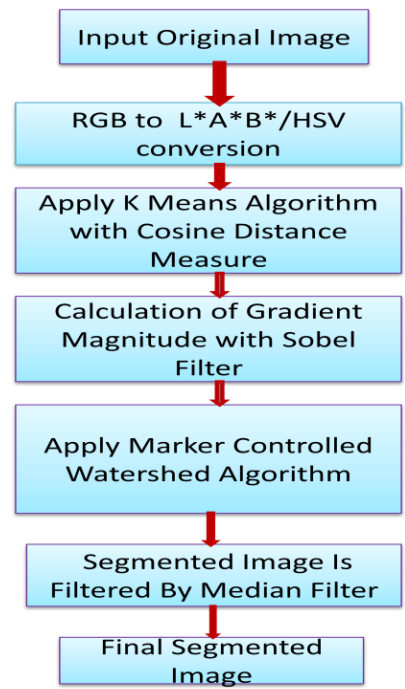

**Fig (VII): Flowchart of Our Color Image Segmentation Approach**

### VI. CLUSTERING & K-MEANS ALGORITHM

Clustering is always recognized as an important area of data mining and has its applications in almost every field of science and engineering. Clustering can be defined as a process of dividing data elements into different groups (known as clusters). But on doing so, one should have to maintain two properties: (1) High intra cluster property and (2) Low inter cluster property. According to first property, elements belong to same cluster should exhibit high similar properties and the second property says that elements of one cluster should be different from the elements of another cluster [34]. Clustering can be classified into two types: (1) Hard Clustering (or Exclusive Clustering) & (2) Soft Clustering (Overlapping Clustering) [35]. In hard clustering, an element strictly belongs to a single cluster only and it cannot share properties with another cluster. While in case of soft clustering, for every element a membership value is maintained and according to that membership value, the element may belong to more than one cluster. Soft clustering gives more accurate results in comparison to hard clustering in some situations [35].

But soft clustering involves more complex computations as fuzzy sets are used to cluster data and fuzzy computation always increases complexity. Also, it is very hard to find out the correct exponent value for the partition matrix in soft clustering depending on the value of which the fuzziness of the algorithm appears [36]. That's why, we have selected hard clustering methodology for our segmentation task and among the hard clustering algorithms, K-Means algorithm is chosen because of its simplicity [35].

The steps involved in K-Means algorithm are:
1. Select an initial partition with k clusters
2. Generate a new partition by assigning each pattern to its closest cluster center.
3. Compute new cluster centers.
4. Continue to do steps 2 and 3 until centroids do not change or memberships finalize.

This algorithm aims at minimizing an *objective function*, in this case a squared error function[37][38]. The objective function is

$$J = \sum_{j=1}^{k}\sum_{i=1}^{n} \left\| x_i^{(j)} - c_j \right\|^2$$

Where $\left\| x_i^{(j)} - c_j \right\|^2$ is a chosen distance measure between a data point $x_i^{(j)}$ and the cluster centre $c_j$, is an indicator of the distance of the *n* data points from their respective cluster centers. K-Means algorithm has very low complexity [35] but only disadvantage with this algorithm is that it is very hard to determine the accurate number of clusters beforehand. A false assumption on the number of clusters may lead to a false clustering result.

### VII. Cosine Distance Metric For K-Means Algorithm

Choosing a right distance metric for K-Means clustering algorithm is a very important issue[39].Cosine distance metric is a measure of similarity between two vectors of n dimensions by finding the cosine of the angle between them. This is based on the measurement of operation and not magnitude [40]. As this distance measure devotes much towards the orientation of data points which is much more related to our work for dealing with pixels of an image, so we have chosen this to be used in K-Means algorithm [8]. It can be defined mathematically as follows : Given an *m*-by-*n* data matrix X, which is treated as *m* (1-by-*n*) row vectors x1, x2,..., x*m*, the cosine distances between the vector x*s* and x*t* are defined as follows[41] :

$$d_{st} = 1 - \frac{x_s x_t'}{\sqrt{(x_s x_s')(x_t x_t')}}$$

### VIII. Sobel Operator

The Sobel operator is generally used in edge detection algorithms to create an image by emphasizing edges and transitions of the image [42]. This is named after Irwin Sobel, the creator of this operator. Actually, this is a discrete differentiation operator which computes an approximation of the gradient of the image intensity function. This is based on convolving the image with a small, separable, and integer valued filter in horizontal and vertical direction and is therefore relatively inexpensive in terms of computations [43]. Actually, this is an orthogonal gradient operator [44], where gradient corresponds to first derivative and gradient operator is a derivative operator. For an image, here involves two kernels: $G_x$ and $G_y$ ; where $G_x$ is estimating the gradient in x-direction while $G_y$ estimating the gradient in y-direction.

Then the absolute gradient magnitude will be given by:

$|G| = \sqrt{(G_x^2 + G_y^2)}$

But often this value is approximated with [43]:

$|G| = | G_x |+| G_y |$

This operator can smooth the effect of random noises in an image. It is differentially separated by two rows and columns, so the edge elements on both sides become enhanced resulting a very bright and thick look of the edges[43]. These are the reasons why we have selected sobel operator for our work [8][15].

### IX. Marker-Controlled Watershed Segmentation

Digabel and Lantuejoul proposed this image segmentation technique[45]. This segmentation algorithm belongs to region based image segmentation algorithms. The geographical concept of watershed is the main base of creation of this algorithm[46][47]. The method consists of the following steps [48]:
1. First of all, compute a segmentation function (this is an image whose dark regions are the objects we are trying to segment).
2. Compute foreground markers which are the connected blobs of pixels within each of the objects.
3. Compute background markers which are pixels that are not part of any object.
4. Then modify the segmentation function so that it only has minima at the foreground and background marker

5. At last, compute the watershed transform of the modified segmentation function.

## X. MEDIAN FILTER

Median filter is a non linear method. This is used to remove noise from images [49]. E.g., salt and pepper noise can be easily reduced by median filter. The median filter can remove noise very effectively while preserving edges- this is one of its most powerful features. This is why median filter is frequently chosen in image segmentation task as we know the importance of the rule played by edges in the same. The working methodology of this filter is: it moves through the image pixel by pixel replacing each value by median value of the neighboring pixels [8][50]. The pattern of the neighbors is called the "window", which slides, pixel by pixel, over the entire image. The median is then calculated by first sorting all the pixel values from the window into numerical order. At last, replace the considering pixel with the median pixel value.

## XI. MSE AND PSNR

The MSE (Mean Squared Error)is the cumulative squared error between the compressed and the original image, whereas PSNR(Peak Signal to Noise Ratio) is a measure of the peak error[51][52][53]. The formula for calculating MSE is as follows [51]:

$$MSE = \frac{1}{MN} \sum_{y=1}^{M} \sum_{x=1}^{N} [I(x,y) - I'(x,y)]^2$$

Where, I(x,y) is the original image, I'(x,y) is its noisy approximated version (which is actually the decompressed image) and M,N are the dimensions of the images. A lower value for MSE implies lesser error.

The formula for PSNR [51] is

$$PSNR = 10 \cdot \log_{10}\left(\frac{MAX_I^2}{MSE}\right)$$

Where, $MAX_I$ is the maximum possible pixel value of the image. A higher value of PSNR implies the ratio of Signal to Noise is higher. So, a higher value of PSNR is always preferred. The 'signal' here is the original image, and the 'noise' is the error in reconstruction.

## XII. EXPERIMENTS & RESULTS

We have chosen Matlab for our experimental study. The comparative study is performed on the peppers image that is available in Matlab demo image [54]. We have taken the k values (number of clusters) as 3 for the K-Means algorithm and the distance metric chosen is "cosine". Following are the results of the experiments:

*(A) Result with respect to L\*A\*B\* Color Space:*

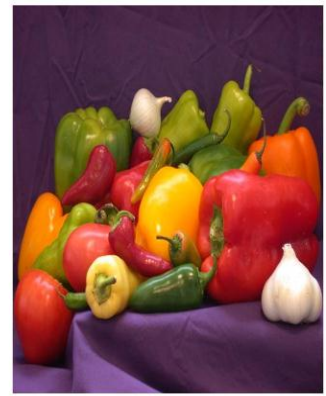

**Fig (VIII): Original Image**

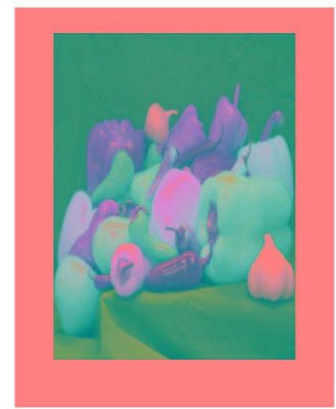

**Fig (IX): L\*A\*B\* Converted Image**

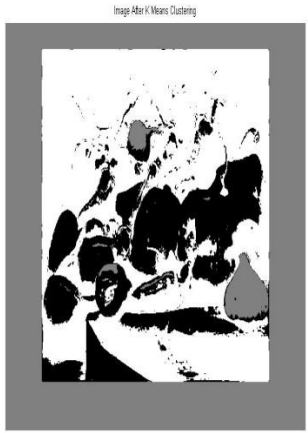

**Fig (X): The Image Obtained After K Means Clustering**

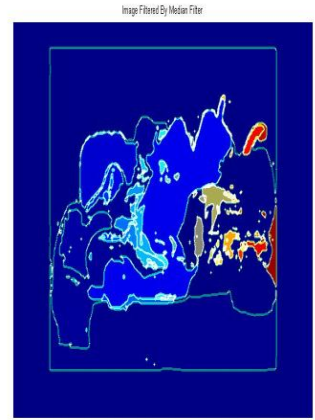

**Fig (XIII): Final Segmented Image Obtained After Applying Median Filter**

*(B) Result with respect to HSV Color Space:*

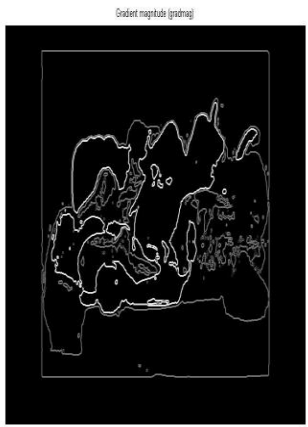

**Fig (XI): Image Obtained After Applying Sobel Filter**

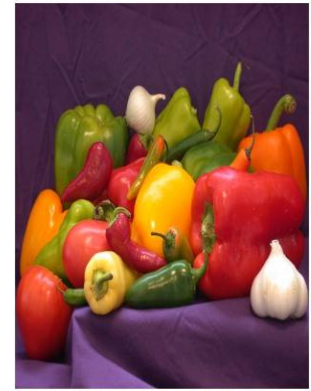

**Fig (XIV): Original Image**

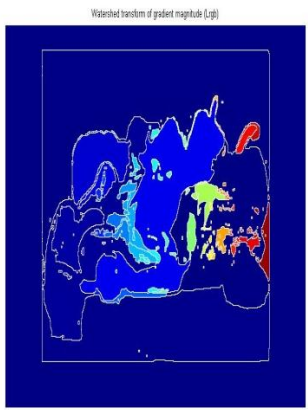

**Fig (XII): Image Obtained After Applying Watershed Algorithm**

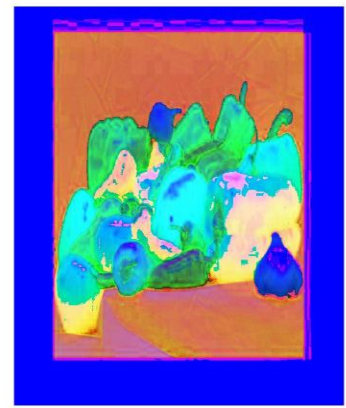

**Fig (XV): HSV Converted Image**

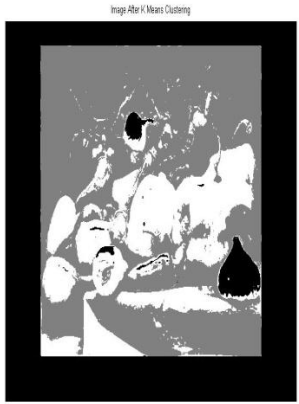

**Fig (XVI): The Image Obtained After K Means Clustering**

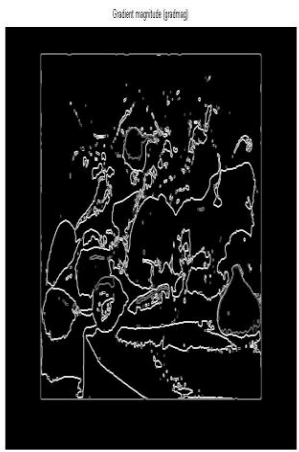

**Fig (XVII): Image Obtained After Applying Sobel Filter**

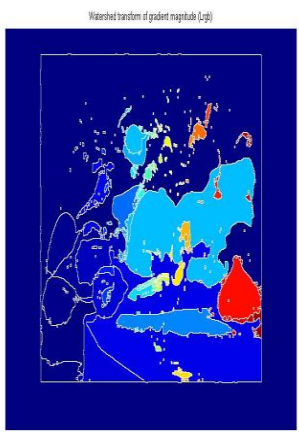

**Fig (XVIII): Image Obtained After Applying Watershed Algorithm**

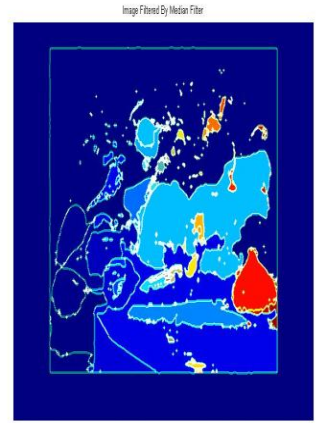

**Fig (XIX): Final Segmented Image Obtained After Applying Median Filter**

*MSE & PSNR Values Calculated With Respect To L*A*B* Color space:*

| MSE : | PSNR : |
|---|---|
| MSE(:,:,1) = | PSNR(:,:,1) = |
| 2.3740e+06 | -15.5901 |
| MSE(:,:,2) = | PSNR(:,:,2) = |
| 4.1124e+05 | -7.9762 |
| MSE(:,:,3) = | PSNR(:,:,3) = |
| 3.2296e+04 | 3.0734 |

*MSE & PSNR Values Calculated With Respect To HSV Color space:*

| MSE : | PSNR : |
|---|---|
| MSE(:,:,1) = | PSNR(:,:,1) = |
| 2.3709e+06 | -15.5844 |
| MSE(:,:,2) = | PSNR(:,:,2) = |
| 4.0902e+05 | -7.9527 |
| MSE(:,:,3) = | PSNR(:,:,3) = |
| 3.1285e+04 | 3.2115 |

*Comparing the Values of MSE & PSNR for L*A*B* & HSV Color spaces:*

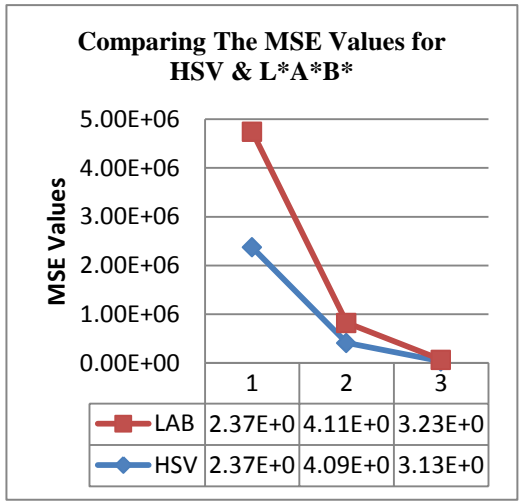

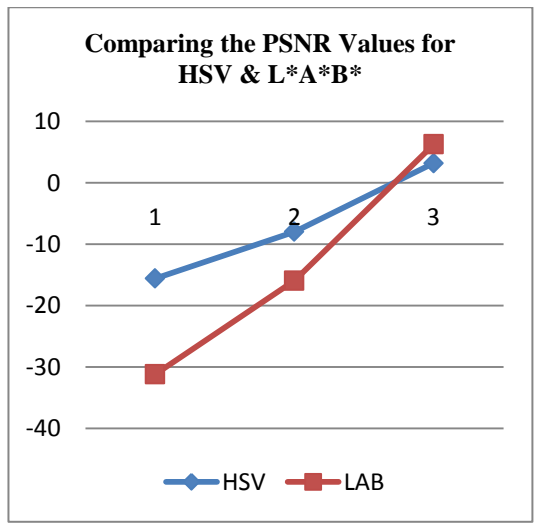

So, from the experimental results, it is observed that HSV Color Space is showing high PSNR values and low MSE values in comparison to that of L*A*B* color space. We know that a high PSNR value and a low MSE value always imply a good segmentation result, hence, we can say that segmentation in HSV color space is showing better performance than the segmentation in L*A*B* color space.

### XIII. CONCLUSION

In today's image processing research, color image segmentation becomes a leading topic for the researchers as because color images allow for more reliable image segmentation than for gray scale images.

Also, the segmented result of color image is more close to the human eyes perceive. But, there exists some parameters upon which the whole process of the segmentation depends. One of such important parameter is the color space. There are different types of color spaces available for use. Each color space carries a few special features with relation to color images. After the image segmentation process, the final segmented image should not have a large amount of noise as otherwise it will hamper the image analysis process. So, we should go for using such color space which has the ability of dealing with noises. Means to say, color spaces should carry as low noise as possible. L*A*B* and HSV are the two frequently chosen color spaces for color image segmentation research. Both are good at dealing with noises and at the top among other color spaces available. In this paper, we have performed a comparative analysis between these two color spaces. MSE and PSNR are taken as the tools for this comparative research. The experimental results show that HSV color space performs better than L*A*B* color space. So, we come to the conclusion that HSV color space will be more suitable for dealing with segmentation of noisy color images. Hence, in our future research, while dealing with noisy image segmentation, then we will give first preference to HSV color space.

### XIV. FUTURE ENHANCEMENT

In future, we will perform an in-depth study on HSV color space. We will try to examine each and every features of HSV color space. Also, it is very important to analyze the properties of this color space keeping an eye on the visual perception of the variation in values for hue, saturation and intensity (H, S and V) for an image pixel of a color image. We can optimize our color image segmentation process if we can understand our color space in a better way. So, we will perform image segmentation with different algorithms for a color image on the HSV color space and try to find out which one will be better suitable with this color space. Currently, we are adopting hard clustering approach for the color image segmentation with HSV color space. In future we will try to extend our work to soft clustering also.